\documentclass[10pt,twocolumn,letterpaper]{article}

\usepackage{wacv}              

\usepackage{graphicx}
\usepackage{amsmath}
\usepackage{amssymb}
\usepackage{booktabs}
\usepackage{xr}
\usepackage{pifont} 
\usepackage{svg}
\usepackage{multirow}
\usepackage{enumitem}

\usepackage[pagebackref,breaklinks,colorlinks]{hyperref}

\def\Dataset{Syned} %

\usepackage[capitalize]{cleveref}
\crefname{section}{Sec.}{Secs.}
\Crefname{section}{Section}{Sections}
\Crefname{table}{Table}{Tables}
\crefname{table}{Tab.}{Tabs.}

\def\wacvPaperID{757} 
\def\confName{WACV}
\def\confYear{2025}

\begin{document}

\title{Reframing Image Difference Captioning with BLIP2IDC and Synthetic Augmentation}

\author{Gautier Evennou$^{1,2}$, Antoine Chaffin${^{3}}^,$\thanks{Work performed while at IMATAG/IRISA.}, Vivien Chappelier$^1$, Ewa Kijak$^2$ \\
$^1$IMATAG, France, $^2$IRISA, CNRS, France, $^3$ LightOn, France   \\
{\tt\small \{gautier, vivien.chappelier\}@imatag.com,antoine.chaffin@lighton.ai, ewa.kijak@irisa.fr}
}
\maketitle

\begin{abstract}

The rise of the generative models quality during the past years enabled the generation of edited variations of images at an important scale.
To counter the harmful effects of such technology, the Image Difference Captioning (IDC) task aims to describe the differences between two images. While this task is successfully handled for simple 3D rendered images, it struggles on real-world images. The reason is twofold: the training data-scarcity, and the difficulty to capture fine-grained differences between complex images.
To address those issues, we propose in this paper a simple yet effective framework to both adapt existing image captioning models to the IDC task and augment IDC datasets. 
We introduce BLIP2IDC, an adaptation of BLIP2 to the IDC task at low 
computational cost, and show it outperforms two-streams approaches by a significant margin on real-world IDC datasets. 
We also propose to use synthetic augmentation to improve the performance of IDC models in an agnostic fashion.
We show that our synthetic augmentation strategy provides high quality data, leading to a challenging new dataset well-suited for IDC named \Dataset\footnote{The code, weights and dataset are available at \url{https://github.com/gautierevn/BLIP2IDC}
 }.



\end{abstract}



\section{Introduction}
\label{sec:intro}
Misinformation, frequently propagated through manipulated or out-of-context images, poses a significant challenge. Image Difference Captioning (IDC) offers a solution by generating textual descriptions that enable humans to readily ascertain whether an image has undergone semantic alterations.
This paper delves into IDC \cite{park2019robust}, a recent approach that goes beyond traditional image analysis by generating detailed textual descriptions of the differences between two images. IDC finds its application across various fields, from detecting subtle changes in satellite imagery \cite{jhamtani2018learning, chouaf2021captioning, qiu2021describing, liu2023progressive} to identifying anomalies in medical imagery \cite{liu2023contrastive} or misinformation explanation.

\begin{figure}
    \centering
    \includegraphics[width=\linewidth]{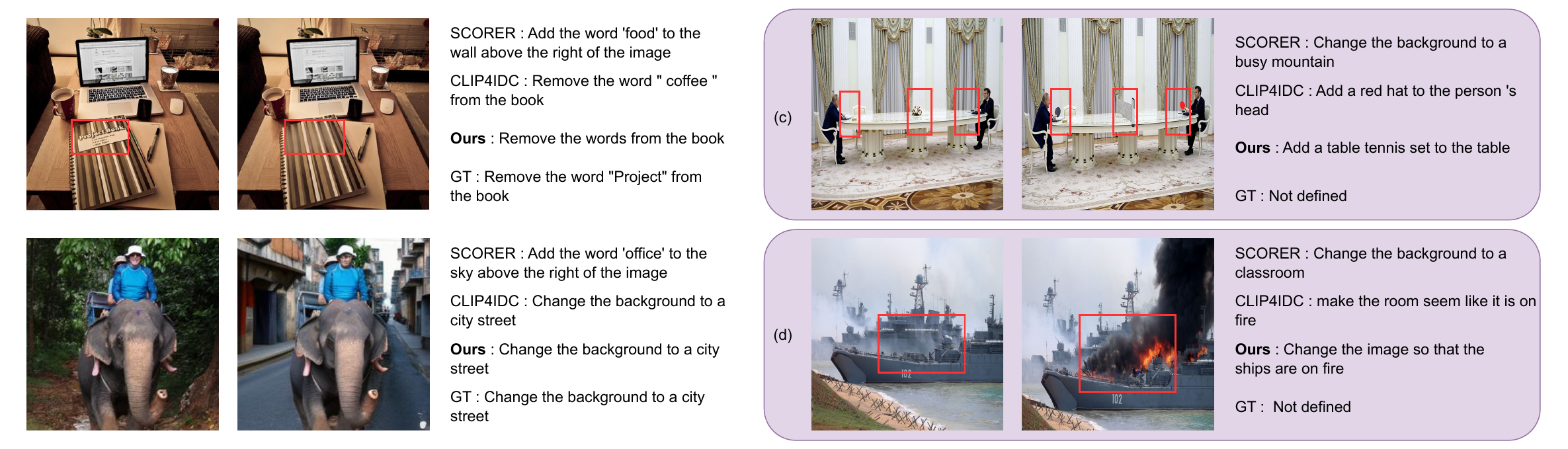}
    \caption{Results of state-of-the-art IDC models on: (a) a train sample of Emu Edit dataset, (b) a test sample and (c-d) zero-shot samples found in the wild. BLIP2IDC (\textbf{Ours}) is able to capture fine-grained differences, describe complex scenes, and generalize well to unseen data. }
    \label{fig:qualitative_analysis}
\end{figure}

Although single-image captioning presents considerable challenges on its own, IDC introduces further complexities, as it involves describing the subtle differences present in a pair of similar images. Ideally, captions should disregard common objects between the images and instead emphasize the nuanced changes between them. 
The advancement of IDC in recent years relies significantly on the emergence of vision-language models and cross-domain learning techniques. 

The second challenge of IDC is the availability of sufficiently large and diverse dataset for the task. 
Creating a high-quality IDC dataset is particularly difficult and resource-intensive, as it requires image pairs with detailed descriptions of their differences, encompassing various types of changes. This process can rely on expensive and time-consuming crowd-sourced labor \cite{Zhang2023MagicBrush}, or alternatively, on the use of 3D rendered scenes \cite{park2019robust} or on capturing temporal variations \cite{tan2019expressing}.
Existing IDC training/evaluation pipeline suffer from shortcomings such as suboptimal metrics, inconsistencies in ground truths, small scale of real-world datasets, or the missing categorization of modifications.

In this paper, we propose an innovative use of synthetic data and advanced multimodal architectures to adress IDC limitations in terms of both data and model. 
We propose the application of generative models \cite{brooks2023instructpix2pix,rombach2022highresolution} to produce synthetic data, providing IDC models with diverse and complex image pairs. 
Concerning the advancements in multimodal models, we discuss how the pre-training of multimodal models like CLIP \cite{guo2022clip4idc} and BLIP2 \cite{li2023blip2} is instrumental for IDC tasks. We explore the potential of BLIP2 in IDC, offering a more 'out-of-the-box' solution contrasting sharply with the previous models reliance on complex, multi-step training and separate processes for image encoding. Despite its larger size and the associated costs for fine-tuning, we demonstrate the feasibility and effectiveness of LoRa \cite{hu2021lora} to adapt BLIP2 to IDC, capitalizing on its extensive model capacity while maintaining a low adaptation cost. The main contributions are~:
\begin{enumerate}[noitemsep,topsep=0pt]
    \item We provide a framework for synthetic augmentation to design well-suited IDC datasets based on diffusion models. We release \textbf{\Dataset}, our synthetic augmented version of Emu Edit~\cite{Sheynin2023EmuEP}, an image editing dataset, to provide a challenging benchmark for the IDC task. We demonstrate the significant improvement brought by this targeted synthetic augmentation.
   \item We propose \textbf{BLIP2IDC}, a new state-of-the-art IDC model based on BLIP2. Extensive experiments conducted on synthetic and real benchmark datasets demonstrate the strong performances of BLIP2IDC in real-world scenarii, and good generalization to unseen content (\cref{fig:qualitative_analysis}). 
   \item We present a comprehensive evaluation of several leading models on a new real IDC dataset based on Emu Edit outputs, contributing to a clearer understanding of the current model capabilities in IDC.
\end{enumerate}


\section{Related Work}
\label{sec:related_works}

\noindent \textbf{Image Captioning and Multimodal Models.} Image difference captioning is closely related to Image Captioning (IC) and Visual Question Answering (VQA). Image Captioning \cite{li2022blip,wang2022ofa,li2023blip2} aims to describe the content of an image with fine-grained captions. IC models are trained on web-scale datasets \cite{lin2015microsoft,flicker30k,young-etal-2014-image} to harness as much visual knowledge as possible.
IC models attempt to connect text and images with various solutions: syntactic trees from image features \cite{mitchell-etal-2012-midge}, RNN \cite{LSTM} decoding of CNN features with \cite{vinyals2016show} or without attention \cite{vinyals2015tell,bahdanau2016neural}, pre-training task using object-based anchors \cite{li2020oscar}, Seq2Seq learning with unified vocabulary for all the linguistic and visual tokens \cite{wang2022ofa} or leveraging additional embeddings to make a bridge between vision tokens and text tokens \cite{li2023blip2}.  
Multimodal models specialized in VQA use cross-modal attention to tune the visual features extraction to the text prompt, enabling interaction and focusing the model on information relevant to the question. While highly efficient \cite{wang2024cogvlm, liu2023visual}, those models are prone to inconsistency due to their prompt dependency \cite{Zhou_2022}. 
\smallskip

\noindent \textbf{Image Difference Captioning (IDC).} IDC focuses on telling apart the discrepancies between two near-identic images. The differences changing the meaning of the image (semantic changes) are the ones the model should focus upon. On the contrary, editiorial changes (non-semantic) such as compression or rescale should be ignored. Prior work struggles with two main hardships: how to represent the difference features and how to gather this specific kind of data. Deep learning based architectures, such as CNN, CLIP-based encoding and RNN are widely used in this topic to learn the features \cite{jhamtani2018learning,shi2020finding,hosseinzadeh2021image,yao2022image,qiu2021describing,chouaf2021captioning,liu2023progressive}.
Previous work extract features for each image independently, thus discarding the images correlation in the pixel space. After this extraction step, they fuse embeddings using concatenation\cite{tu2023selfsupervised, yao2022image,qiu2021describing, park2019robust} before feeding them to whether a difference encoder for change representation, or directly to a decoder to generate the text descriptions. The difference encoder and decoder are transformer-based \cite{vaswani2017attention} or RNN-based \cite{qiu2021describing}.
Another axis of work is to define pretraining tasks to find a suitable representation space for the differences \cite{tu2023selfsupervised}. Recently, the use of a multimodal model such as CLIP \cite{radford2021learning} enables for a better representation, thus achieving state-of-the-art performances. Yet, current adaptation to IDC \cite{guo2022clip4idc} struggles with exploiting all its pretraining. Moreover, because high-quality data is hard to find, most methods are trained and evaluated on 3D rendered datasets, that do not accurately represent real-world performance.                                                                                                                                                                                                                                                    
\smallskip

\noindent \textbf{Image Editing.} The recent progress in generative models \cite{goodfellow2014generative, rombach2022highresolution} enables realistic generation of images.  This further propell work \cite{meng2022sdedit,brooks2023instructpix2pix,parmar2023zeroshot} on image editing, where semantic modifications are performed to an image using a textual prompt, while keeping the main subject of an image.
Recent works \cite{Sheynin2023EmuEP,Zhang2023MagicBrush} focused on using already existing real-world text-image datasets such as MSCOCO \cite{lin2015microsoft} to perform editions on those images, by building on modifications of the associated caption. The main hurdle in these methods is the fully supervised generation and verification process, where each generation is reviewed by operators and several level of filters are used to keep the best generations. While this process produces a high quality dataset, it is time-consuming and costly. 
We show in this paper that such setup nevertheless allows the automatic creation of IDC datasets: a couple is created using the original image and its modified version, and the prompt used to perform the edit is used as the target modification description.

\section{BLIP2IDC}
\label{sec:blip2idc}

BLIP2 \cite{li2023blip2} is a multimodal model that introduces the Querying Transformer (QFormer) as the main block to connect text and image representations. It relies on a two-stage pretraining: a vision-language representation learning stage with a frozen image encoder, and a vision-to-language generative learning stage with a frozen LLM. 
The QFormer is composed of two transformers modules. The first one attends to the image through cross attention between learned embeddings, called queries, and vision encoder outputs. The second module interacts with the ground truth text and with the queries embeddings to ensure vision-language alignment. BLIP2 is pre-trained to jointly optimize three pre-training objectives that are Image-Text Contrastive Learning, Image-Text Matching, and Image-grounded Text Generation.
As a contribution we show that, taking advantage of its pretraining for Image Captioning, BLIP2 can be adapted to IDC at a low computational cost and without changing its architecture.

\subsection{Adaptation}
\begin{figure}
    \centering
    \includegraphics[width=\linewidth]{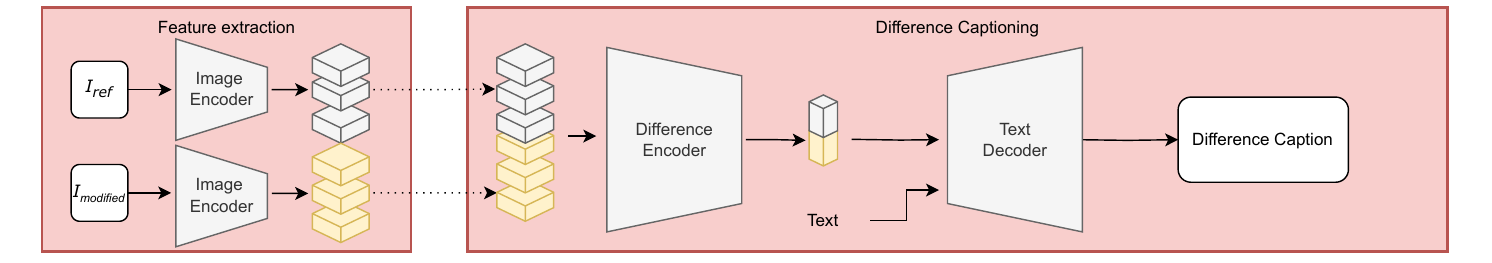}
    \caption{Classic IDC pipeline. The feature extraction step is performed before training. The training step leverages the features from a frozen image encoder by concatenating them and then learns the difference representation from these concatenated features. The output is then fed to the text decoder to perform the generation of the difference caption in an autoregressive manner.} 
    \label{fig:classic_idc_pipeline}
\end{figure}

\begin{figure}
    \centering
    \includegraphics[width=\linewidth]{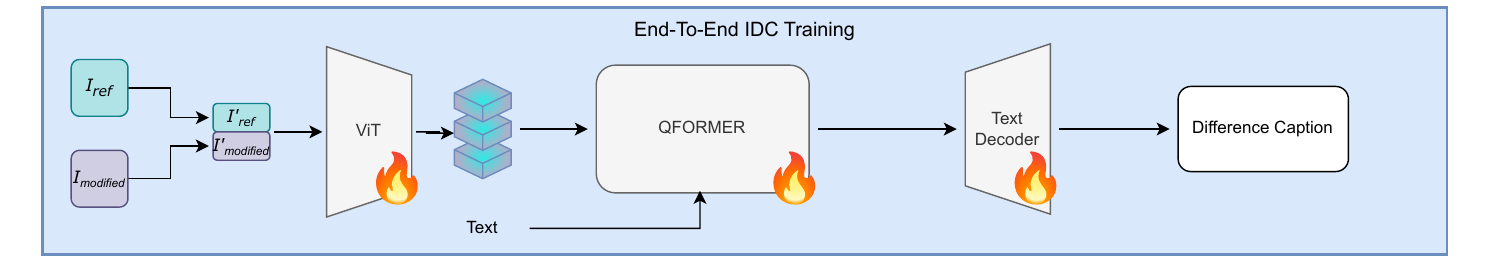}
    \caption{BLIP2IDC end-to-end pipeline.  We first resize and concatenate the image inputs before feeding them to the BLIP2 architecture fine-tuned using LoRA.} 
    \label{fig:blip4idc_pipeline}
\end{figure}

We argue that the classic IDC two-streams encoding scheme featured in \cref{fig:classic_idc_pipeline} results in a suboptimal comparison of images at the features level.
For standard image captioning, BLIP2 takes one image as input and uses a ViT model \cite{dosovitskiy2021image,vaswani2017attention} as a frozen image encoder. 
In the context of IDC, feeding BLIP2 jointly with the two images to be compared enables the attention layers of the visual encoder and QFormer to focus early on the differences between the 2 images, and to encode the two images in relation to each other rather than separately. 
Our BLIP2 adaptation pipeline is shown in \cref{fig:blip4idc_pipeline}. We give a single image as input, resulting from the vertical concatenation of two images, allowing the model to attend to the differences early, while avoiding any modification of the architecture of the model. Unlike BLIP2, which only trains the QFormer, the ViT and the LLM should be fine-tuned to compensate for this modification of the input domain.

Although images are stretched by concatenation, it does not undermine the performances.
We conjecture this behaviour is due to the way BLIP2 was pretrained, where images are randomly cropped then stretched to a fixed square size, without padding. Thus, the BLIP2 model learned to be robust to various stretching operations.\par

\subsection{Efficient fine-tuning}
\label{sec:finetuning_blip2}

\noindent \textbf{Modules fine-tuning.}
Although BLIP2IDC does not require to change the architecture of the model, the model should be fine-tuned for the IDC task to adapt to the new task and type of data. Unlike the original training of BLIP2 which only fine-tunes the QFormer, all the components, including the ViT, QFormer and image-grounded text decoder, are finetuned to get the best IDC performance (see \cref{fig:ablation_study_lora} in the result section).


\smallskip
\noindent \textbf{Low Rank Adaptation (LoRA).} We use Low Rank Adaptation (LoRA), to reduce training resources while maintaining performance. By fine-tuning just 0.1\% of all parameters—specifically, the attention modules' $Q, K, V$ layers—we achieve top performance with minimal resource use. This approach allows BLIP2\-IDC to adapt to IDC tasks, balancing high performance with resource efficiency.


\subsection{Advantages over existing IDC models}
\label{sec:idc_comparison}

The latest leading models for IDC are CLIP4IDC \cite{guo2022clip4idc} and SCORER \cite{tu2023selfsupervised}. CLIP4\-IDC utilizes a two-step training approach, first to adapt image-pairs visual representations for captioning and then to generate captions based on visual differences encoding, through contrastive and cross-entropy losses respectively. SCORER follows the traditional IDC framework as depicted in \cref{fig:classic_idc_pipeline}, enhancing performance with innovative modules for view-shift invariance and caption informativeness, and is presented as showing particular strength on the CLEVR-DC dataset.


These models are characterized by complex training procedures with several steps and separate image encoding processes. In contrast, our model BLIP2IDC, simply concatenates images for joint encoding  employing early attention mechanisms while learning in only one step. This approach proves crucial for IDC, as evidenced by our experimental results in \cref{sec:experiments}. 
BLIP2IDC also benefits from knowledge gained through the large-scale unsupervised pre-training of BLIP2 for image captioning, making it compelling given the challenges associated with training an IDC model from scratch on fewer data.


\section{Datasets}

The structure of IDC datasets is a triplet $(I_{ref}, I_{modified}, {GT})$ which respectively represents the original image $I_{ref}$, the modified image $I_{modified}$, and the ground-truth set of reference descriptions of the differences $GT$. 
We compare our approach to recent IDC models on standard IDC datasets: CLEVR-Change \cite{park2019robust}, CLEVR-DC \cite{Kim_2021_ICCV}, Spot-The-Diff \cite{qiu2021describing} and Image Editing Request \cite{tan2019expressing}. 
These datasets are distinguished by various properties, summarized in \cref{tab:idc_datasets}: the number of image pairs ($I_{ref}, I_{modified}$), the origin of the $I_{ref}$ images, which may be real images or generated by 3D models, the number of reference captions (ground-truths $GT$) per triplet, the way in which $I_{modified}$ and $GT$ are obtained, the number and type of transformations performed, and whether some manual intervention was required to create the dataset. This may happen at the level of triplet filtering or the creation of ground-truth legends.

These properties can limit the creation of these datasets, which can be costly, as well as their usefulness, for example when the diversity of transformation types is reduced. We discuss the limitations of these datasets hereafter, as well as the issues posed by the evaluation of the IDC task. To handle those limitations, we propose a method based on text-guided image editing methods for creating IDC datasets without manual intervention, which can then be adapted to different use cases.
Furthermore, instruction-based image editing capabilities are evaluated on some benchmarks (IE datasets), like MagicBrush \cite{Zhang2023MagicBrush} or Emu Edit test set generations (EE) \cite{Sheynin2023EmuEP}. Those datasets are also composed of triplets $(I_{ref}, T_{instruction}, I_{edited})$, with $T_{instruction}$ the modification that maps $I_{ref}$ to $I_{edited}$ and $I_{edited}$ the edited image. We thus propose to adapt those datasets for the IDC task.

\begin{table*}[ht]
\caption{IDC and IE datasets. Setting indicates the origin of reference images as 3D scenes or from the real world. Editing method refers to the way modified images are created or collected. GT is the number of ground-truths in the validation and test set.} 
\centering
\begin{tabular}{ l| c| l| l| c| c} \hline 
\centering
Dataset & \# image pairs &  \ setting & editing method & curated & \# GT \\ \hline 
CLEVR-Change \cite{park2019robust} & 79,606 & 3D Scenes & 3D engine & $\checkmark$ & 5.0 \\  
CLEVR-DC \cite{Kim_2021_ICCV} & 48,000 & 3D Scenes & 3D engine & $\checkmark$ & 5.0 \\  
STD \cite{qiu2021describing} & 13,192 & Real-world & Temporal change & & 1.86 \\ 
IER \cite{tan2019expressing} & 3,939 & Real-world & Human Editing & $\checkmark$ & 3.0\\ \hline
MagicBrush \cite{Zhang2023MagicBrush} & 10,308 & Real-world & DallE-2 \cite{ramesh2022hierarchical} platform &  $\checkmark$ & 1.0\\ 
EE \cite{Sheynin2023EmuEP} & 5,612 & Real-world & Automatic generation &  $\checkmark$ & 1.0 \\
\Dataset (ours) & 28,720 & Real-world & Automatic generation & & 5.0  \\ \hline
\end{tabular}
\label{tab:idc_datasets}
\end{table*}

\smallskip
\subsection{Existing datasets}

\noindent \textbf{CLEVR-Change.} Introduced by Robust Change Captioning \cite{park2019robust}, the goal of this dataset is to evaluate primary abilities for visual understanding such as being able to tell the shape, the color, the material and the position of an object and if one or none of those properties changed. This dataset enables to assess if an IDC model understands spatial relationships and if it is robust to non-semantic changes. Five types of scene changes are defined. Editing instruction and ground-thruths captions are automatically constructed following a template. 
However, this dataset was created with a 3D engine, and all the elements are 3D scenes far from the real world setting, leading to a model that doesn't transfer very well to real-life examples.\par
\smallskip
\noindent \textbf{CLEVR-DC.}
CLEVR-DC \cite{Kim_2021_ICCV} is a derivative of CLEVR-Change with extreme shift in viewpoints. It uses the same types of modifications as CLEVR-Change. The goal of this dataset is to evaluate the view-invariancy of the IDC model but it shares the limitations of the original dataset. 

\smallskip
\noindent \textbf{Spot-The-Diff (STD).} Unlike CLEVR-Change, STD \cite{jhamtani2018learning} uses a real world setting for the pairs of images. To cope with the issue of producing modified images, it uses temporal evolution. Thus, a pair of images is made of one image taken at a given time and another of the same place taken at a later time, with a fixed point of view. Most of the images are set in a parking lot, heavily restricting the diversity of ground-truth as most changes in a parking are add/remove people/car. Although heavily biased, this dataset enables one to test IDC models on real-world images. \par
\smallskip
\noindent \textbf{Image Editing Request (IER).} Born from crawling Reddit and Zhopped, the IER  dataset \cite{tan2019expressing} relies on instructions from online users in Photoshop specialized forums. The input image and editing instruction were posted, and people sent their modified images on Reddit. Crawling those results of the manual edition led to a really high quality and diverse dataset but with a really small scale. Each image pair of the test set has three GT captions, written by three different annotators. 
IER provides a wide range of semantic modifications, as for example "add a sailor hat to ducks", "replace the background by a spaceship", or "replaces brooms with lightsabers".\par

\smallskip
\noindent \textbf{MagicBrush.} This dataset is the first large-scale, manually-annotated instruction guided image editing dataset \cite{Zhang2023MagicBrush} covering various global and local editing scenarii. MagicBrush comprises 10K $(I_{ref}, T_{instruction}, I_{edited})$ triplets, which is sufficient to fine-tune large-scale image editing models with the use of LoRA. For this dataset, modifications are not grouped into categories like in CLEVR-Change, making it hard to explain performance and target specific weaknesses.

\smallskip
\noindent \textbf{Emu Edit (EE).} This dataset \cite{Sheynin2023EmuEP} from Meta is composed of pairs of original images from MSCOCO \cite{lin2015microsoft} and modified images generated by the Emu Edit model, a state-of-the-art generation model, which takes an image and an editing instruction as inputs. 
This dataset also contains fine-grained information about the type of modifications generated, grouped in eight different categories~: Add, Text, Color, Background, Local Style, Global and Remove. Paired with the real-world setting of original images which enables for greater usability, these properties make it a great candidate for a challenging IDC benchmark. 

\subsection{IDC task issues}

IDC models are trained on triplets $(I_{ref}, I_{modified}, GT)$ and evaluated on a test set of pairs $(I_{ref}, I_{modified})$ by comparing the generated caption with the reference captions $GT$, using various automatic evaluation metrics, such as BLEU, ROUGE, METEOR and CIDEr~\cite{BLEU,ROUGE, METEOR,CIDER}. Some aspects of the real-world datasets weaken the quality of the evaluation.

\smallskip
\noindent \textbf{Suboptimal metrics.} The references-based metrics more accurately evaluate
the quality of generated sentences if multiple references are provided. Five references is usually the minimum used. However, none of the avalaible real-world datasets provides a consistent and sufficient number of references, making evaluations less accurate. 

\smallskip
\noindent \textbf{Lack of consistency in ground truths.} In STD and IER, the humanly-annotated ground-truth captions do not consistently refer to the same difference or set of differences, especially when several modifications are made to the image. Some captions describe a single modification, others several. An IDC model trained to describe only the main difference will struggle with this type of ground truth, and it is the case for all existing IDC models. Either all the differences should be mentionned in each ground-truth, or just one, consistently, to avoid conflicting objectives.

\smallskip
\noindent \textbf{Tiny real-world datasets.} As real-world datasets are very time-consuming to make, they are small in size, with at most 10,000 text-image pairs. As a result, they can lead to models specialized in specific subsets of possible modifications, as for STD. On the other side, when modifications are very diverse, as for IER, a type of modification can occur only few times. Metrics then heavily depend on the data splits, according to whether a modification appears only in the training or in the test set. This is all the more possible if the modifications are not grouped into classes allowing stratified sampling.

\smallskip
\noindent \textbf{Modifications categorization.} Grouping changes into categories is useful for stratified sampling, but also for analyzing results, as it makes it easier to identify difficult changes and the weaknesses of IDC models. 


\subsection{Synthetic augmentation}
\label{sec:aug}

\begin{figure}
    \centering
    \includegraphics[width=\linewidth]{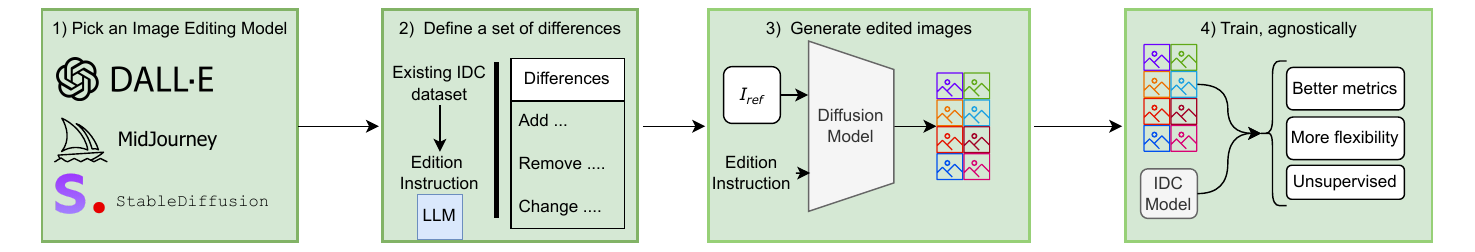}
    \caption{Synthetic dataset creation pipeline leveraging a prompt-based image editing and large language models. }
    \label{fig:synthetic_pipeline}
\end{figure}

To cope with the current shortcomings of IDC datasets mentioned above, we propose a pipeline to generate synthetic training samples based on real-world original images. The global pipeline is illustrated in \cref{fig:synthetic_pipeline}. The idea is to leverage the emerging prompt-based image editing models to produce modified images based on a set of original images and editing instructions belonging to a defined range of modifications. Additional GT captions are generated by a Large Language Model (LLM) as variations of the editing instructions, ensuring consistent GTs.\par

Such a pipeline can be used either to augment an existing IDC dataset, or to create a new dataset whose images and modifications are tailored to a particular downstream task with new specific types of data or modifications.

To evaluate this pipeline, we applied it to augment the EE dataset \cite{Sheynin2023EmuEP}, as this dataset provides real-world edited images alongside editing instructions well-suited for an image editing model and with a clear distinction between each class of modifications. As image editing model, we choose the InstructPix2Pix \cite{brooks2023instructpix2pix} model fine-tuned on the MagicBrush \cite{Zhang2023MagicBrush} dataset, the state-of-the-art solution for executing real-world edits. The train set is augmented by generating eight new modified images per original ones from the EE dataset and their associated editing instructions, without any manual curation. 

\begin{figure}
    \centering
    \includegraphics[width=0.6\linewidth]{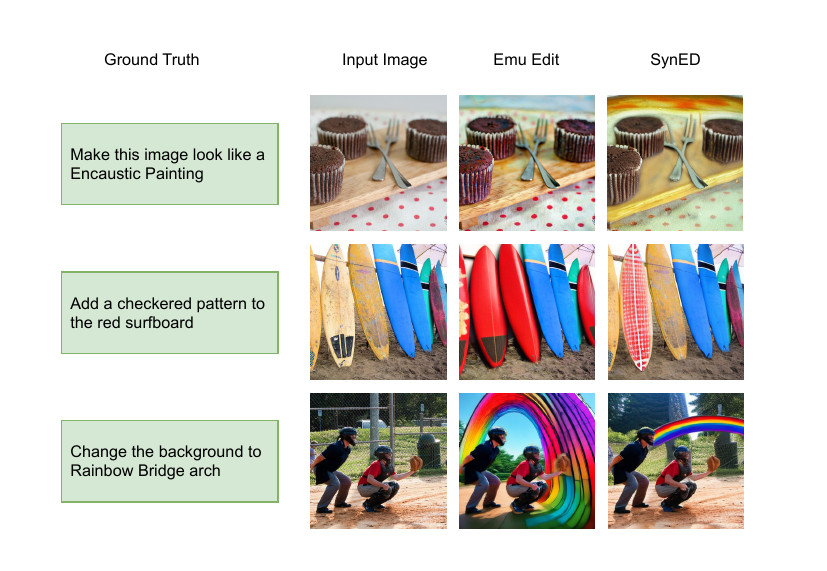}
    \caption{Comparison of train set samples from EE, generated by the Emu Edit model, and \Dataset,~generated by a fine-tuned InstructPix2Pix model. }
    \label{fig:syned}
\end{figure}

To ensure meaningful evaluation, we leverage the \href{https://huggingface.co/meta-llama/Llama-2-7b-chat-hf}{Llama-2-7b-chat-hf} model to generate 4 additional variations of each editing instructions from the EE test set (cf. supplementary). Each of the 2,022 samples in the test set is therefore described by 5 GT captions: the original instruction and the 4 reference captions provided by the LLM. Note that the train set images (see \cref{fig:syned}) were not modified by the same editing model as in the test set. 

The resulting \Dataset~dataset comprises :
\begin{itemize}[noitemsep,topsep=0pt]
    \item \textbf{a train set of 28,720 pairs of $(I_{ref}, I_{modified})$}: 8 variations of each of the 3,590 original images from MSCOCO, generated by InstructPix2Pix. The different modifications are categorized in 8 classes. 
    \item \textbf{a test set of 2,022 samples with 5 references each}, where the modified images were generated by the Emu Edit model. 
\end{itemize}

 Although the modifications can be defined as anything according to the downstream needs, they have to be aligned with the image editing model's capabilities. For instance, editing models currently fall short in adding text to images. Such instructions introduce bad samples in the training set. However, as image editing models improve, the proposed pipeline will allow to include more reliable edits to the training set of IDC models, although it already allows for customized generation of more challenging IDC datasets. 

\section{Experiments}
\label{sec:experiments}

\subsection{Evaluation protocol}

\noindent \textbf{Models and datasets.} We compare BLIP2IDC to other state-of-the-art IDC models with the results reported in their respective paper on the four standard IDC datasets: CLEVR-Change, CLEVR-DC, STD, and IER. 
We also introduce the EE dataset to the IDC task, on which we also train and evaluate CLIP4IDC \cite{guo2022clip4idc} and SCORER \cite{tu2023selfsupervised} for the sake of comparison.

\smallskip
\noindent \textbf{Finetuning.}

BLIP2IDC is fine-tuned as described in \cref{sec:finetuning_blip2}, with a LoRA of rank $8$. For our own adaptation of BLIP2, we use the decoder-only BLIP2's LLM version, with \href{https://huggingface.co/facebook/opt-2.7b}{opt2.7B} and \href{https://huggingface.co/google/vit-base-patch16-224}{vit-base-patch16-224} as the ViT model. We use LoRA implementation from the \href{https://github.com/huggingface/peft}{peft}\cite{peft} python library to train BLIP2 and to store adaptation weights in a lightweight 20 MB file. For \href{https://github.com/sushizixin/CLIP4IDC}{CLIP4IDC} and 
\href{https://github.com/tuyunbin/SCORER}{SCORER}, we either used the hyper-parameters from their respective online repository, or hyperparameters communicated by the authors (see supplementary). 


\smallskip
\noindent \textbf{Standard data augmentation.} To increase the robustness of the model, we use RandomGaussianBlur and JPEG Compression as non-disruptive data augmentation scheme. Augmentations should indeed leave the ground-truth semantic content untouched. Transformations based on crop, color jitter, horizontal/vertical flip, brightness or contrast can add meaningful differences and thus change the expected difference caption.

\smallskip
\noindent \textbf{Metrics.}
Based on previous work \cite{jhamtani2018learning,shi2020finding,hosseinzadeh2021image,yao2022image,qiu2021describing,chouaf2021captioning,liu2023progressive}, CIDEr (C) is the main metric used to evaluate the generated difference captions. This n-gram-based approach ensures that the evaluated text captures the most relevant aspects of the image as agreed upon by multiple human annotators. In a semantic-dependent context, the CIDEr metric rewards descriptions that accurately reflect the consensus understanding of the image content. 
BLEU-4 (B), ROUGE-L (R) and METEOR (M) are also used as secondary metrics to assess the sentences quality.

\begin{table*}
\caption{CIDEr scores on CLEVR-Change for the 5 categories of changes.} 
\centering
\begin{tabular}{l c c c c c}
\hline
Method & Color & Texture & Move & Add & Drop  \\
\hline
DUDA \cite{park2019robust} (ICCV 2019) & 120.4 & 86.7 & 56.4 & 108.2 & 103.4 \\
VAM+  \cite{VAM} (ECCV 2020) & 122.1 & 98.7 & 82.0 & 126.3 & 115.8\\
IFDC \cite{IFDC}(TMM 2022)& 133.2 & 99.1 & 82.1 & 128.2 & 118.5\\
DUDA+ \cite{hosseinzadeh2021image} (CVPR 2021) & 120.8 & 89.9 & 62.1 & 119.8 & 123.4\\
BiDiff \cite{BDLSCR} (IJIS 2022) & 115.9 & 106.8 & 71.8 & 121.3 & 124.9\\
IDC-PCL \cite{yao2022image} (AAAI 2022) & 131.2 & 101.1 & 81.7 & \underline{133.3} & 116.5\\
CLIP4IDC \cite{guo2022clip4idc} (ACL-IJCNLP 2022) & \underline{149.1} & \underline{135.3} & 91.0 & 132.4 & \textbf{135.5}\\
SCORER \cite{tu2023selfsupervised} (ICCV 2023) & 143.2 & 135.2 & 91.6 & 129.4 & 132.6\\
SCORER+CBR \cite{tu2023selfsupervised}(ICCV 2023) & 146.2 & 133.7 & \underline{92.2} & 131.1 & \underline{133.9} \\
\hline
BLIP2IDC (Ours) & \textbf{152.31} & \textbf{137.0} &	\textbf{99.8} & \textbf{135.6} & 133.1\\
\hline

\end{tabular}
\label{tab:sem_chang_cc}
\end{table*}
\subsection{BLIP2IDC evaluation}
We evaluate the performance of BLIP2IDC against other methods on the four existing IDC datasets. All results except CLIP4IDC on CLEVR-DC and BLIP2IDC are reported from previous work.

\smallskip
\noindent \textbf{Results on 3D scenes.}
Results per type of semantic changes on CLEVR-Change are given in ~\cref{tab:sem_chang_cc}. 
BLIP2IDC ranks first on almost all types of semantic changes, with a 10\% increase in the captioning performance on the most difficult 'Move' change, and a significant improvement over previous state-of-the-art method on the CIDEr metric. 
On CLEVR-DC, that introduces extreme viewpoint changes, we observe that CLIP4IDC achieves the best results, followed by BLIP2IDC (\cref{tab:combined_results}). Both models are better than SCORER although the latter was specially designed to ensure extreme-view-shift invariancy.

\smallskip
\noindent \textbf{Results on real-world images.}
According to Table ~\ref{tab:combined_results}, the BLIP2IDC method ranks first on the CIDEr metric by a significant 9.2\% margin with respect to state-of-the-art on STD.
The IER dataset is made of photoshopped images in a real world setting. Thus, the impressive improvement over previous state-of-the-art is expected since BLIP2 was pretrained on 129M images in a real-world setting \cite{lin2015microsoft,Krishna2016VisualGC, sharma-etal-2018-conceptual}. Our results shows that our adaptation outperforms state-of-the-art models, even those based on CLIP which was trained at large scale too.
Overall, those results show that efficiently adapting a powerful pretrained model yields state-of-the-art performance, whereas most existing models fall behind due to the lack of data during training.


\begin{table*}[h]
\centering
\caption{Results on CLEVR-DC, STD, and IER, reported from their corresponding papers, except $^\dagger$:  reported from our own experiments.}
\begin{tabular}{l c c c | c c c | c c c c} 
\hline
 & \multicolumn{3}{c|}{CLEVR-DC} & \multicolumn{3}{c|}{STD} & \multicolumn{4}{c}{IER} \\
\textbf{Method} & B & M & C & B & M & C & B & M & R & C \\ \hline
Dyn rel-att \cite{tan2019expressing} & - & - & - & - & - & - & 6.7 & 12.8 & 37.5 & 26.4 \\
M-VAM \cite{VAM} & 40.9 & 27.1 & 60.1 & 10.1 & 12.4 & 38.1 & - & - & - & - \\ 
M-VAM+RAF \cite{VAM} & - & - & - & 11.1 & 12.9 & 43.5 & - & - & - & - \\ 
VA \cite{Kim_2021_ICCV} & 44.5 & 29.2 & 70.0 & - & - & - & - & - & - & - \\
VACC \cite{Kim_2021_ICCV} & 45.0 & 29.3 & 71.7 & 9.7 & 12.6 & 41.5 & - & - & - & - \\ 

DUDA \cite{park2019robust} & 40.3 & 27.1 & 56.7 & 8.1 & 12.5 & 34.5 & 6.5 & 12.4 & 37.3 & 22.8 \\ 

NCT \cite{NCT} & 47.5 & 32.5 & 76.9 & - & - & - & 8.1 & \underline{15.0} & 38.8 & 34.2 \\ 

SRDRL+AVS \cite{SRDRL} & - & - & - & - & 13.0 & 35.3 & - & - & - & - \\
IFDC \cite{IFDC} & - & - & - & 8.7 & 11.7 & 37.0 & - & - & - & - \\
BDLSCR \cite{BDLSCR} & - & - & - & 6.6 & 10.6 & 42.2 & 6.9 & 14.6 & 38.5 & 27.7 \\
VARD-Trans \cite{Vard} & 48.3 & 32.4 & 77.6 & - & 12.5 & 30.3 & \underline{10.0} & 14.8 & 39.0 & \underline{35.7} \\ 
MCCFormers-D \cite{qiu2021describing} & 46.9 & 31.7 & 71.6 & 10.0 & 12.4 & 43.1 & 8.3 & 14.3 & 39.2 & 30.2 \\ 
SCORER \cite{tu2023selfsupervised} & \underline{49.5} & \textbf{33.4} & 82.4 & 9.4 & \underline{13.8} & 38.5 & 9.6 & 14.6 & 39.5 & 31.0 \\ 
SCORER+CBR \cite{tu2023selfsupervised} & 49.4 & \textbf{33.4} & 83.7 & 10.2 & 12.2 & 38.9 & \underline{10.0} & \underline{15.0} & 39.6 & 33.4 \\ 
CLIP4IDC \cite{guo2022clip4idc} & \textbf{54.7}$^\dagger$  & \underline{33.0}$^\dagger$ & \textbf{89.9}$^\dagger$ & \textbf{11.6} & \textbf{14.2} & \underline{47.4} & 8.2 & 14.6 & \underline{40.4} & 32.2 \\ 
\hline
BLIP2IDC (Ours)$^\dagger$ & 49.3 & \underline{33.0} & \underline{88.5} & \underline{11.4} & 13.5 & \textbf{51.4} & \textbf{17.4} & \textbf{20.1} & \textbf{48.5} & \textbf{74.1} \\ \hline
\end{tabular}
\label{tab:combined_results}
\end{table*}

\begin{figure}
    \centering
    \begin{subfigure}{0.30\textwidth}
        \includegraphics[width=\linewidth]{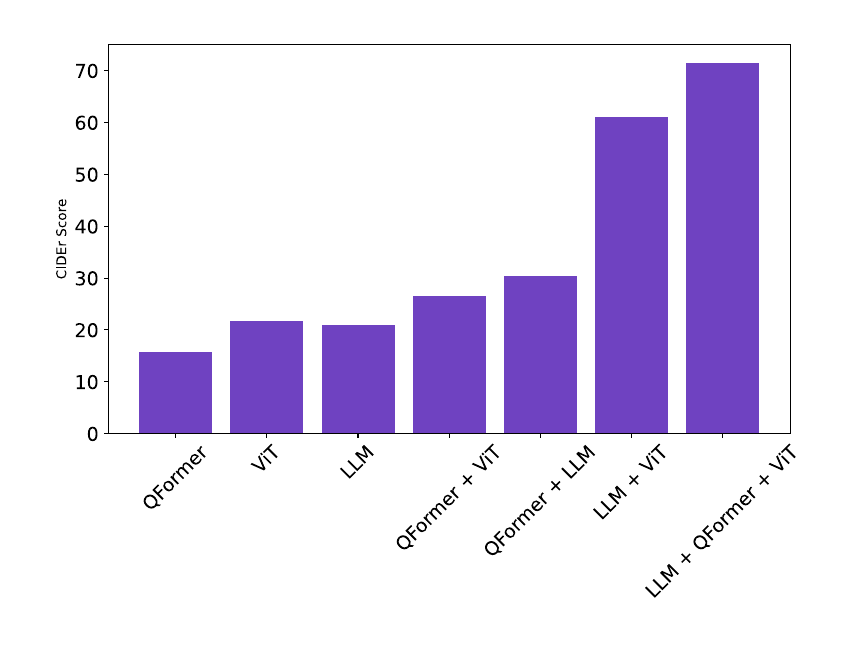}
        \caption{IER score}
    \end{subfigure}
    \hfill 
    \begin{subfigure}{0.30\textwidth}
        \includegraphics[width=\linewidth]{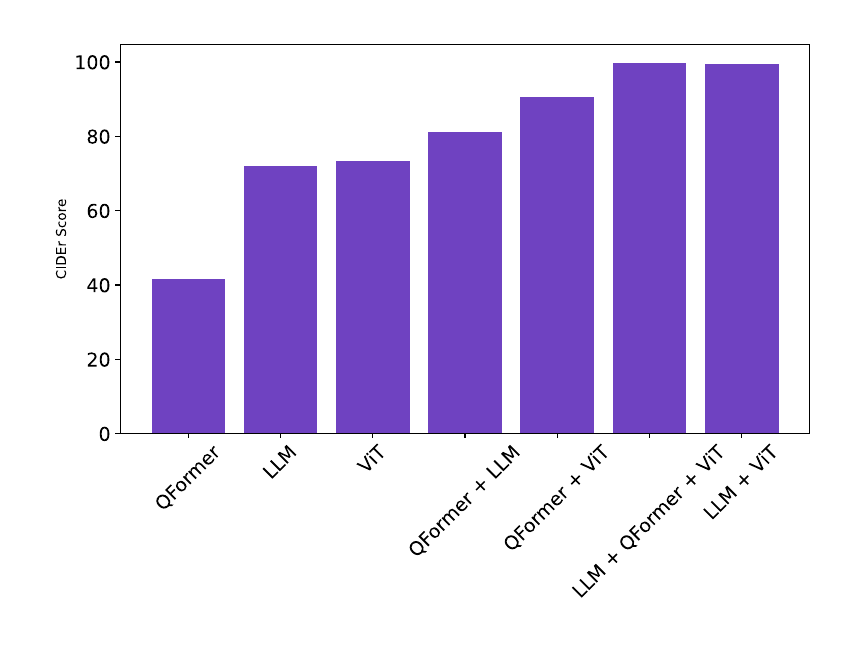}
        \caption{EE Score}
    \end{subfigure}
        \hfill 
    \begin{subfigure}{0.30\textwidth}
        \includegraphics[width=\linewidth]{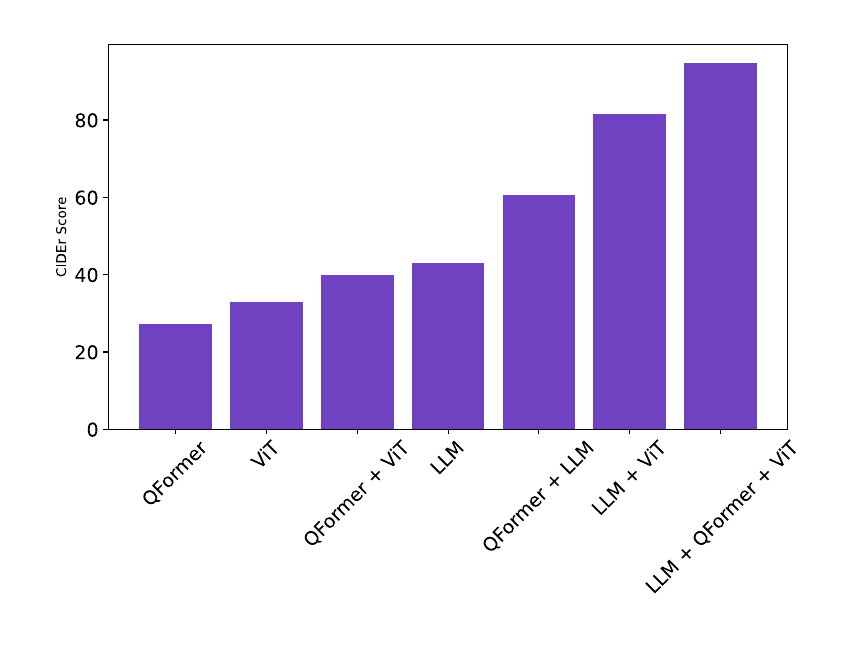}
        \caption{\Dataset~Score}
    \end{subfigure}
    \caption{Ablation Study of BLIP2IDC's CIDEr according to fine-tuned modules on different datasets.}
    \label{fig:ablation_study_lora}
\end{figure}

\smallskip
\noindent \textbf{Ablation study.}
The performance variations across different BLIP2IDC configurations, depending on the fine-tuned modules, are depicted in~\cref{fig:ablation_study_lora}. This analysis highlights the significant impact of fine-tuning both the ViT and LLM modules. Given the mismatch between the ViT and LLM modules for both the input and output domains, the QFormer not surprisingly stands out as the less critical component for fine-tuning, as it already had undergone extensive training during the BLIP2 pretraining phase. 

\subsection{Automatically generated dataset}

We introduced in \cref{sec:aug} \Dataset,~a new dataset designed for the IDC task, built upon the EE dataset. New data are generated by a different instruction-based image editing model, 
thus broadening the scope and diversity of the new modified images wrt the original EE dataset, as depicted in \cref{fig:syned}. We evaluate the usefulness of Syned as data augmentation for different IDC models in \cref{tab:Synthetic_augmentation_performance}. 
We show an improvement with the proposed synthetic augmentation (EE + \Dataset) for both CLIP4IDC and BLIP2IDC. Note that the available implementation of SCORER enables to reach at most a CIDEr score of 23.2  which is not significant in this setting and can be achieved only by reward hacking. 


\begin{table}[ht]
\centering
\caption{CIDEr scores on Emu Edit test set. We show consistent improvements with synthetic augmentation (EE + \Dataset) on all models. }
\begin{tabular}{p{2cm}ccc}
\hline
\textbf{Models} & \textbf{EE} &  \textbf{\Dataset + EE  }  & Improvement (\%) \\
\hline
SCORER & 21.1 & 23.2 & -\\
CLIP4IDC & 32.4 & 35.9 & 10.8 \\
BLIP2IDC      & 100.0 &  106.8 & 6.8 \\
\hline
\end{tabular}
\label{tab:Synthetic_augmentation_performance}
\end{table}



\begin{table*}[ht]
\centering
\caption{CIDEr Performance Metrics by Change Category and Dataset Version. This table lists the CIDEr score of each following categories : Add (A), Text (T), Background (B), Color (C), Style (S), Global (G), Remove (R), Local (L) and Overall (O) which is the CIDEr computed over all the samples of the test set.}
\label{tab:performance_metrics_emu}
\begin{tabular}{p{2cm}ccccccccc}
\hline
Dataset & A & T & B & C & S & G & R & L & O \\ \hline
\Dataset & \underline{102.45} & 147.17 & 111.78 & 151.87 & 12.77 & \textbf{31.56} & 72.27 & 61.00 & 94.83 \\
EE & 101.01 & \textbf{147.69} & \textbf{119.48} & \underline{155.76} & \underline{30.53} & 29.46 & \underline{93.39} & \underline{76.24} & \underline{100.83} \\
\Dataset + EE & \textbf{107.38} & \underline{147.42} & \underline{112.37} & \textbf{164.40} & \textbf{37.07} & \underline{30.43} & \textbf{111.16} & \textbf{77.81} & \textbf{106.83} \\ \hline
\end{tabular}
\end{table*}

In Table \ref{tab:performance_metrics_emu}, we study BLIP2IDC behaviour with respect to its training data, being either \Dataset, EE or both on different modification categories. 
Combining datasets is generally the best configuration overall, except for the text (T) category. We assume this is due to the poor capacity of most Image Editing models to generate text in images, which hampers the performance of the synthetic augmentation. The worst categories in terms of CIDEr score performance are the Style and Global categories, which are very subjective changes. More details are available in the supplementary material.

\subsection{Qualitative analysis}

We present in \cref{fig:qualitative_analysis} a comparative analysis of state-of-the-art IDC models outputs across different scenarii: in-distribution image-pairs from the train and test set of EE, and out-of-distribution images from the web, to assess the zero-shot capabilities of the models.

We observe SCORER's limitations in handling real-world images in all cases. It only succeed in identifying a text modification in the train sample, and confuses actions like adding or removing.
CLIP4IDC performs well on samples from EE but struggles with zero-shot generalization. Specifically, it misidentifies the "table tennis set" due to its absence in the training data, confusing it with a similar but incorrect object.
Additionnaly, it fails to understand the context in which the object is added.
BLIP2IDC, on the other hand, excels on data in-distribution and demonstrates superior capability in generalization through effective use of cross-attention. It accurately recognizes and describes the entire scenes in~\cref{fig:qualitative_analysis}(c-d), including complex additions like a table tennis set, by leveraging its pretraining. This allows BLIP2IDC to not only identify the presence of fire in~\cref{fig:qualitative_analysis}(d) but also specify the object affected, showcasing its ability to connect semantic changes with their context.

\subsection{Limitations}

BLIP2 is pre-trained on large image-text datasets sourced from the web, such as LAION \cite{schuhmann2022laion5b}. As BLIP2IDC is a finetuned version of BLIP2, it inherits the biases observed on the BLIP2 model due to its pretraining data.
Our synthetic augmentation pipeline also relies on several synthetic generation models and thus inherits their limitations, such as ethnics misrepresentation or adversarial vulnerabilities. The LLM used for generating the ground truth variations will hallucinate, even with careful prompting. \par
Furthermore, Image Editing models, while improving, are still not able to perform every type of modifications. Careless use of these models can deteriorate the dataset quality if the editing instructions are not well aligned with the model capacity.
We finally emphasize the in-domain limitation: IDC datasets can hardly be exhaustive in their ability to handle all types of change, due to the multiplicity of modification interpretations and descriptions. For a given domain, a more suitable approach is to curate a limited number of well defined changes and to build upon them, allowing for better control.

\section{Conclusion}
In this paper, we propose a novel framework that adapt existing image captioning model to the IDC task in order to benefit from their extensive pre-training. 
Image captioning pretraining data are indeed easier to obtain than the one for IDC. This allows to effectively leverage the global world knowledge from large-scale unlabeled dataset and transfer it to IDC using a smaller amount of data.
As a demonstration, we adapt BLIP2 to the IDC task. By innovatively encoding image differences at the pixel level rather than relying on the traditional dual-stream scheme, we ensure a more informative and direct approach to understanding image variations. BLIP2IDC allows to outperform existing methods on standard benchmarks. Additionnaly, we introduce a synthetic augmentation strategy that not only adresses critical challenges of data scarcity and the need for robust, real-world applicable architectures but also sets a new benchmark in IDC performance. This paves the way to the application of IDC to more diverse data and types of edits.


{\small
\bibliographystyle{ieee_fullname}
\bibliography{egbib_test}
}

\end{document}